\documentclass{article}

\usepackage{arxiv}

\usepackage[utf8]{inputenc} 
\usepackage[T1]{fontenc}    
\usepackage{hyperref}       
\usepackage{url}            
\usepackage{booktabs}       
\usepackage{amsfonts}       
\usepackage{nicefrac}       
\usepackage{microtype}      
\usepackage{lipsum}		
\usepackage{graphicx}
\usepackage[numbers,sort&compress]{natbib}
\usepackage{doi}
\usepackage{float}
\usepackage{amsmath}
\usepackage{amssymb}
\usepackage{threeparttable}

\title{New Boolean satisfiability problem heuristic strategy: Minimal Positive Negative Product Strategy}

\author{Zhao Qun\\
	School of Information Resource Management,\\
	Renmin University of China,\\
	Beijing, China, 100872 \\
	\texttt{zhao\_qun@ruc.edu.cn} \\
	\And
    Wang Xintao\\
    Institute of Economics,\\
    Chinese Academy of Social Sciences,\\
    Beijing, China, 100872 \\
    \texttt{wangxintao1583@163.com} \\
    \And
	Yang Menghui \thanks{Corresponding Author}\\
	School of Information Resource Management,\\
	Renmin University of China,\\
	Beijing, China, 100872 \\
    Key Laboratory of Data Engineering and Knowledge Engineering, \\
    Ministry of Education, \\
    Beijing, China, 100872 \\
	\texttt{yangmenghui@ruc.edu.cn} \\
}

\date{}

\hypersetup{
pdftitle={New Boolean satisfiability problem heuristic strategy: Minimal Positive Negative Product Strategy},
pdfauthor={Zhao Qun, Wang Xintao, Yang Menghui},
pdfkeywords={Boolean satisfiability problem \and Conflict-driven clause learning \and heuristic algorithms}
}

\begin{document}
\maketitle

\begin{abstract}
This study presents a novel heuristic algorithm called the "Minimal Positive Negative Product Strategy" to guide the CDCL algorithm in solving the Boolean satisfiability problem. It provides a mathematical explanation for the superiority of this algorithm over widely used heuristics such as the Dynamic Largest Individual Sum (DLIS) and the Variable State Independent Decaying Sum (VSIDS). Experimental results further confirm the effectiveness of this heuristic strategy in problem-solving.
\end{abstract}


\section{Introduction}
\label{sec:Introduction}
The Boolean satisfiability problem (sometimes called propositional satisfiability problem and abbreviated SATISFIABILITY, SAT or B-SAT) is the problem of determining if there exists an interpretation that satisfies a given boolean formula. Boolean Satisfiability is probably the most studied of combinatorial optimization/search problems.This decision problem is of central importance in many areas of computer science, including theoretical computer science, complexity theory\cite{karp2010reducibility}, cryptography\cite{mironov2006applications} and artificial intelligence\cite{vizel2015boolean}.

Conflict-driven clause learning (CDCL) solvers have proven their effectiveness in solving the SAT\citep{davis1960computing,davis1962machine}. CDCL is a mechanism that improves the search of the solution space by learning from past mistakes and avoiding them in future attempts. CDCL solver learns new clauses with conflict information. This allows the solver to trim the search space and prevent redundant exploration of similar areas.

Improvements in the CDCL algorithm start with the application of heuristic algorithms. The choice of a heuristic algorithm is to reach conflicts as soon as possible. While the Dynamic Largest Individual Sum (DLIS) heuristic is known for quickly reaching conflicts\cite{marques1999grasp}, it was criticized for lack of regarding the usefulness of new learned clauses. Chaff introduced the Variable State Independent Decaying Sum (VSIDS) heuristics\cite{marques1999grasp}, which prioritizes variables that have been frequently used in recent conflict analysis.

Chaff achieved significant success, and many subsequent research aimed at improving its performance\citep{pipatsrisawat2007lightweight,huang2007effect}. Other research on the CDCL solvers diverts attention to memory management, some modern CDCL solvers pay attention to removing learned clauses deemed unhelpful. Glucose introduced an approach to assess the value of learned clauses to remove low-value ones\cite{audemard2009predicting}. CaDiCaL and Kissat sometimes discard nearly all learned clauses\cite{balyo2020proceedings}. \citet{kruger2022too} found that a large number of redundant learned clauses can significantly reduce the efficiency of the CDCL solver. This explains why memory management can expedite SAT problem-solving by removing surplus clauses. Furthermore, this research offers a new angle on experimental design. If a heuristic algorithm learns fewer clauses compared to others during problem-solving, it clearly indicates improved efficiency.

This study shifts the focus from clause removal back to proposing a heuristic strategy superior to Chaff's VSIDS. Firstly, the study introduces a straightforward method Positive Negative Product (PN product) to assess the complexity of SAT problems and validates its effectiveness. Secondly, it discusses how to improve DLIS and VSIDS by building upon the PN product. Finally, it demonstrates the effectiveness of the new heuristic strategy Minimal Positive Negative Product Strategy through experimentation. When solving SAT problems of similar complexity, it learns fewer clauses, affirming the success of this improvement.

\section{Related works}
\label{sec:Related works}

Conflict-driven clause learning works as follows: (1) Select a variable and assign True or False. This is called decision state. Remember the assignment. (2)  Apply Boolean constraint propagation (unit propagation). The meaning of unit propagation is explained as follows: IF an unsatisfied clause has all but one of its literals or variables evaluated at False, then the free literal must be True in order for the clause to be True. For example, if the below unsatisfied clause is evaluated with $A = False$ and $B = False$, we must have $C = True$ in order for the clause $A \vee B \vee C $ to be true. (3) Build the implication graph. (4) If there is any conflict: 1. Find the cut in the implication graph that led to the conflict. 2. Derive a new clause which is the negation of the assignments that led to the conflict. 3. Non-chronologically backtrack to the appropriate decision level, where the first-assigned variable involved in the conflict was assigned. (5) Otherwise continue from step 1 until all variable values are assigned.

Dynamic largest individual sum (DLIS) heuristic means that in decision state selects the literal that appears most frequently in unresolved clauses and assigns it. The changes in variable assign and learned clauses evolve unresolved clauses. Frequency needs recalculations, this process is referred to as "dynamic."

Compared with DLIS, Variable State Independent Decaying Sum (VSIDS) heuristic emphasizes paying more attention to new clauses learned in newly triggered conflicts. Variable State Independent Decaying Sum (VSIDS) is described as follows: (1) Each variable in each polarity has a counter, initialized to 0. (2) When a clause is added to the database, the counter associated with each literal in the clause is incremented. (3) The (unassigned) variable and polarity with the highest counter are chosen at each decision state. (4) Ties are broken randomly by default, although this is configurable. (5) Periodically, all the counters are divided by a constant. The fifth step of VSIDS decays relative to the importance of the previously learned clause according to the distance from the current conflict, the divided constant can be regarded as a decay factor that determines the speed of decay.

Maximum Occurrences on Minimum sized clauses (MOM) \cite{freeman1995improvements} is a heuristic algorithm that inspires our Minimal Positive Negative Product Strategy. For the first time, MOM proposed using variables to select alterals based on the frequency of two polarities. DLIS and VSIDS both select variables based only on the variable with the highest frequency of a single polarity. The MOM is as eq. \ref{eq:eq1}, where $f(x)$ is polarity $x$'s frequency in unresolved smallest clauses, $f(\neg x)$ is polarity $\neg x$'s frequency in unresolved smallest clauses, $k$ is chosen heuristically. What needs to be emphasized here is MOM The calculated frequency is in the unresolved smallest clauses, not all unresolved clauses used by DLIS, VSIDS and our strategy.

\begin{equation}\label{eq:eq1}
	((f(x) + f(\neg x)) \cdot 2^k + f(x) \cdot f(\neg x) \tag{1}
\end{equation}

\section{Preliminaries}
A propositional logic formula, also called Boolean expression, is built from variables, operators AND (conjunction, also denoted by $\land$), OR (disjunction, $\vee$), NOT (negation, $\neg$), and parentheses. A formula is said to be satisfiable if it can be made TRUE by assigning appropriate logical values (i.e. TRUE, FALSE) to its variables. The Boolean satisfiability problem (SAT) is, given a formula, to check whether it is satisfiable.

The logic formula of a SAT question study is like eq. \ref{eq:eq2}, which is composed of a large number of clauses, and clauses are composed of literals, and literals are composed of all variables and their two polarities. variable $X$ has two polarities, one is $x$ and the other is $\neg x $. Solving the SAT problem is to assign values to all variables $X$ through the logic formula, and select a polarity for them to be true, so that the logic formula is true. If there is no such assignment scheme, this problem is unsolvable. SAT problems in which each sentence contains K literals are called K-SAT problems.

\begin{align*} \label{eq:eq2}
(¬x_1 \vee ¬x_2 )\land(¬x_1 \vee x_3)\land(¬x_3 \vee ¬x_4 )\land(x_2 \vee x_4 \vee x_5 )\land(¬x_5 \vee x_6\vee¬x_7)\land(x_2 \vee x_7 \vee x_8)\land \\
        (¬x_8 \vee ¬x_9 ) \land(¬x_8 \vee x_{10} )\land(x_9 \vee ¬x _{10}  \vee x_{11})\land(¬x_{10} \vee ¬x_{12} )\land(¬x_{11} \vee x_{12} ) \tag{2}
\end{align*}

\section{Method}

We present a method for describing the complexity of SAT problems, referred to as the "positive negative product," abbreviated as PN product. The calculation of the PN product involves selecting the polarity of variables in all unresolved clauses of the SAT problem. The polarity that appears more frequently is designated as the positive group, and that appears less frequently is designated to the negative group. For instance, if the polarity $\neg x$ appears more often than $x$ in the unresolved clauses, $\neg x$ is included in the positive group. The polarities within the positive and negative groups are counted in all unresolved clauses, and the counts are summed as P and N. The PN product results from multiplying P and N together. We assert that a higher initial PN product in SAT problems signifies greater complexity and greater difficulty in finding a solution.

\begin{equation}\label{eq:eq3}
	PN\_product = P \cdot N \tag{3}
\end{equation}

\begin{figure}
	\centering
	\includegraphics[width=0.80\textwidth]{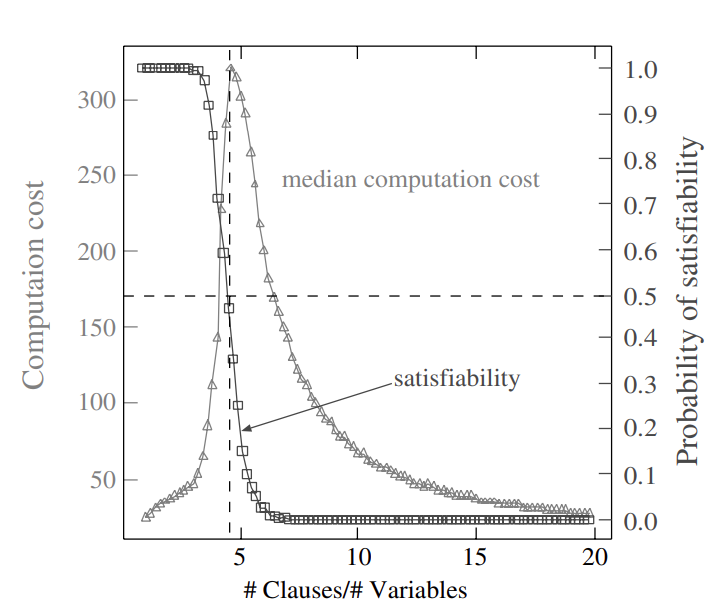}
	\caption{The phase transition of 3-SAT problems}
	\label{fig:fig1}
\end{figure}

The PN product provides a straightforward way to describe the difficulty level of an SAT problem. A phase transition exists in the ratio of variables to clauses in the Boolean satisfiability problem. For 3-SAT problems, the phase transition point is at 4.26, which means that when the ratio of the number of clauses to variables is 4.26, it is most challenging to determine whether the problem has a satisfying solution. Fig \ref{fig:fig1}\cite{zhang2001phase}illustrates the phase transition in 3-SAT problems. The horizontal axis represents the ratio of the number of clauses to the number of variables, and the vertical axis shows the computational complexity and the likelihood of having a solution. It is observed that the probability of finding a solution rapidly decreases when the ratio of the number of clauses to variables is between 4 and 6, and the computational complexity for solving problems is highest in this range.

\newpage

We designed an experiment to generate 10,000 sets of 3-SAT problems with 100 variables and 426 clauses. The experiment found that whether it is solvable is significantly negatively correlated with the initial PN product of the problem as table \ref{tab:table1}.

\begin{table}[ht]
    \caption{OLS Regression Test Result of solvable and initial PN product}
    \centering
    \begin{tabular}{lllllll}
        \toprule
        ~ & \multicolumn{6}{l}{solvable} \\
        \cmidrule(r){2-7}
        ~ & coef & std err & t & P>|t| & [0.025 & 0.975]\\
        Intercept & 22.3404 & 0.593 & 37.695 & 0.000 &21.179 &23.502 \\
        initial PN product & -5.63e-05 & 1.53e-06 & -36.883 & 0.000 &-5.93e-05 &-5.33e-05\\
        \bottomrule
    \end{tabular}
    \label{tab:table1}
\end{table}

\begin{figure}
	\centering
	\includegraphics[width=0.80\textwidth]{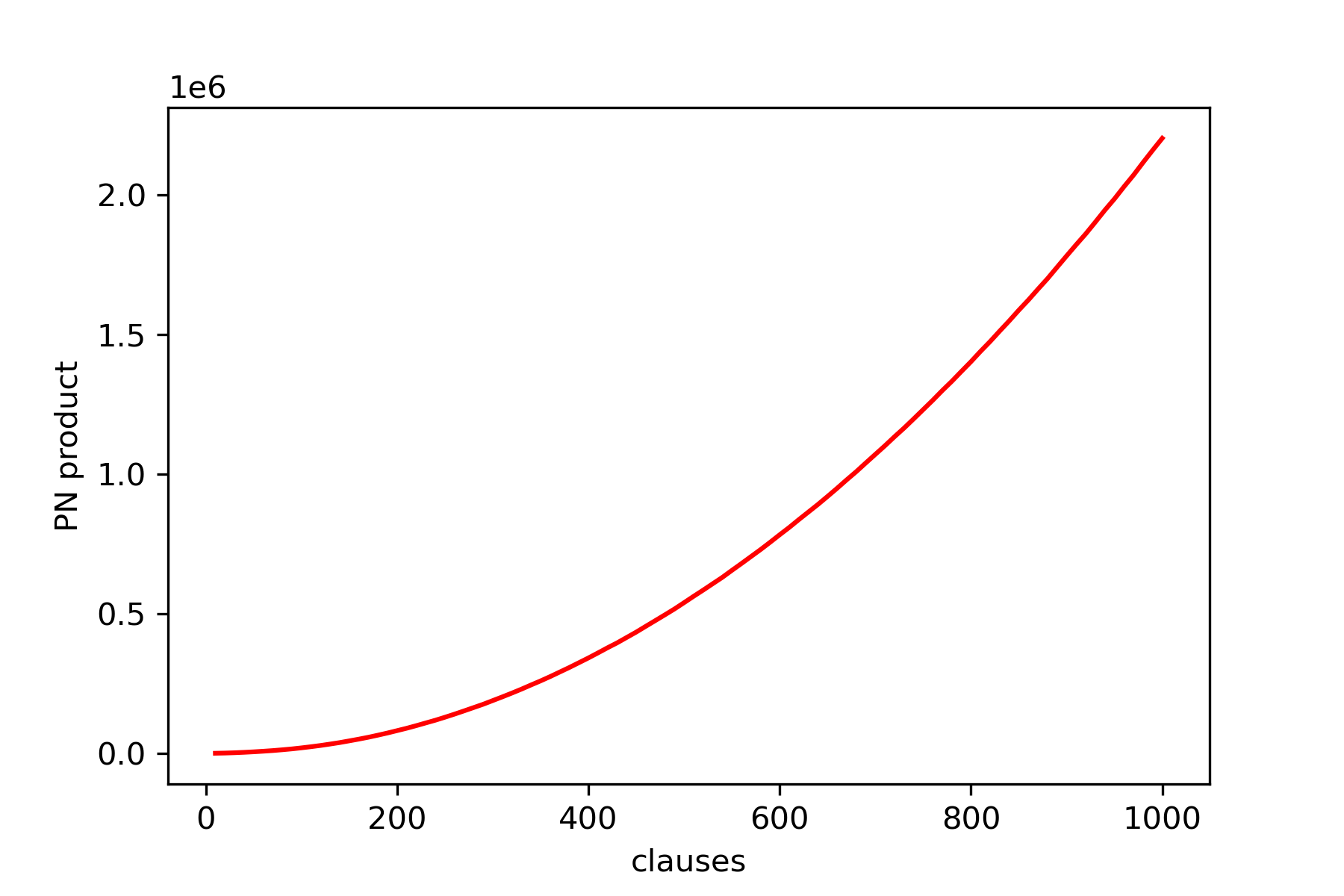}
	\caption{Initial PN product increases monotonically with clauses}
	\label{fig:fig2}
\end{figure}

When the number of variables remains constant, an SAT problem becomes increasingly unsolvable as the number of clauses increases. The PN product increases as more clauses, show less likely to find a solution. In a 3-SAT problem with 100 variables, Fig \ref{fig:fig2} shows the more clauses, the higher the initial PN product becomes. To solve SAT problems, the SAT solver tries to reduce the PN product of unresolved clauses to zero by assigning values. If can't bring the PN product to zero, it means no solution. With a fixed number of variables, the more clauses added, the harder reduce the PN product to zero. A high initial PN product indicates an unsolvable problem.

\newpage

The CDCL solver also attempts to reduce the PN product to zero through variable assignment. When the CDCL solver encounters a conflict and learns clauses, it causes the PN product value of unresolved clauses to increase. If the increased PN product cannot be brought to zero, it is determined no suitable solution for the problem. In the decision state, choosing a variable $X$ and assigning it the polarity from the positive group, $p$ represents the frequency of variable $X$ in the unresolved clauses with the positive group's polarity. $n$ represents the frequency of variable $X$ in the unresolved clauses with the negative group's polarity. $p \cdot (k-1)\cdot \frac{P}{P + N}$ and $p \cdot (k-1)\cdot \frac{N}{P + N}$ denote the approximate count of polarities. Those polarities are not assigned but no longer belong to unresolved clauses because the belonged clauses are resolved. $k$ represents the average number of literals in clauses. After $X$ assignment, new PN product as eq. \ref{eq:eq3}.If unit propagation occurs, it will trigger new variable assignments, leading to a further decrease in the PN product. Using a heuristic algorithm to select a variable for assignment, may cause a chain reaction, resulting in a series of variables being automatically assigned.

\begin{equation}\label{eq:eq3}
	new\_PN\_product = (P-p-p \cdot (k-1)\cdot \frac{P}{P + N})\cdot(N-n-p \cdot (k-1)\cdot \frac{N}{P + N}) \tag{3}
\end{equation}

DLIS and VSIDS heuristic algorithms accelerate the process of reducing the PN product by selecting variables for assignment. Speeding up this process can lead to faster solution discovery or declaring the problem as unsolvable. To understand why DLIS and VSIDS heuristic strategies need improvement, Although VSIDS introduces a decay factor, it fundamentally operates similarly to DLIS by choosing the polarity that appears most frequently in unresolved clauses for assignment, focusing on maximizing $p$ while ignoring the impact of $n$. As shown in eq. \ref{eq:eq3}, maximizing $p$ and $n$ at the same time contributes to a quicker reduction of the PN product. On one hand, selecting larger $n$ variables for assignment in the decision state results in a more significant decrease in the PN product. On the other hand, selecting larger $n$ variables also accelerates unit propagation because it assigns more literals in unresolved clauses to FALSE, forcing more variables in the remaining clauses to be assigned TRUE. A better heuristic algorithm should consider maximizing both $p$ and $n$.  We recommend selecting variables for assignment with relatively large values of $p$ and $n$ as a new heuristic strategy. Subsequent experiments will test the effectiveness of the $p + n$ and $p \cdot n$ approaches.

\section{Results}

The main way to assess SAT algorithms is through computation time. However, this time is influenced by programming languages and hardware. In this study, we evaluate SAT algorithms based on the final number of clauses in the formula. When the CDCL algorithm faces conflicts, it adds learning clauses to the formula. More clauses in the formula mean longer solving times, and an overly long formula increases the time needed for searches and slows down computations. This is why many studies focus on removing learned clauses.

We used DLIS and VSIDS as baselines for comparison. We designed heuristic algorithms using both $p+n$ and $p \cdot n$ to select variables without decay factors. We also combined them based on the MOM approach, such as $(p + n)\cdot 4 + p \cdot n$ and $(p + n)\cdot 32 + p \cdot n$. We also included a decay factor in the $p \cdot n$ heuristic algorithm. For the experiments, we selected 100 variables and 426 clauses in 3-SAT problems, conducting 500 repetitions and recording the maximum, average, and median clause counts after resolution. The experimental results are shown in the table \ref{tab:table2}, where smaller maximum, average, and median are better.

\begin{table}[H]
    \centering
    \begin{tabular}{p{30pt}p{30pt}p{30pt}p{30pt}p{70pt}p{70pt}p{30pt}p{70pt}}
    \toprule
        ~ & DLIS & VSIDS & $p + n$ & $(p + n) \cdot 32 + p \cdot n$ & $(p + n) \cdot 4 + p \cdot n$ & $p \cdot n$ & $p \cdot n$ with decaying \\
    \cmidrule(r){1-8}
        maximum & 47296 & 28916 & 15230 & 10485 & 10204 & 12384 & 9426 \\
        average & 3781.624 & 3899.068 & 2437.302 & 2391.58 & 2306.94 & 2290.548 & 2191.346 \\ 
        median & 2467.0 & 2648.5 & 2037.5 & 2037.5 & 1944.0 & 1834.0 & 1778.0 \\
    \end{tabular}
    \label{tab:table2}
\end{table}

Firstly, from the experimental results, it is evident that both $p + n$ and $p \cdot n$ algorithms, which consider $n$ perform better overall than DLIS and VSIDS, which only consider $p$.  Secondly, the performance of the $p \cdot n$ algorithm is superior to that of the $p + n$ algorithm. Lastly, the $p \cdot n$ algorithm with a decay factor outperforms all others.

\section{Conclusion}

This study introduces a method called the "positive negative product" to simply measure the complexity of K-SAT problems, and experimental results validate the effectiveness of this metric. Based on the mechanism of the positive negative product, the study points out where the two mainstream CDCL heuristic algorithms, DLIS and VSIDS, can be improved and demonstrates the effectiveness of these enhancements. In the experiments, it is observed that the introduction of the negative group polarity count $n$ in the $p \cdot n$ heuristic is superior to the $p + n$ approach. Moreover, with the addition of the VSIDS decay factor, the heuristic algorithm based on $p \cdot n$ achieves the best performance. This heuristic strategy is referred to as the "Minimal Positive Negative Product Strategy." A limitation of this study is that it does not explore why $p \cdot n$ yields better results than $p + n$.

\bibliographystyle{unsrtnat}
\bibliography{references}  






\end{document}